%% file: main.tex
\documentclass{article} % For LaTeX2e
\usepackage[preprint]{colm2026_conference}

\usepackage{microtype}
\usepackage{hyperref}
\usepackage{url}
\usepackage{booktabs}

\usepackage{lineno}

\definecolor{darkblue}{rgb}{0, 0, 0.5}
\hypersetup{colorlinks=true, citecolor=darkblue, linkcolor=darkblue, urlcolor=darkblue}

\title{Selecting Feature Interactions for Generalized Additive Models by Distilling Foundation Models}

\author{
Jingyun Jia$^{1}$ \quad Chandan Singh$^{2}$ \quad 
Rich Caruana$^{3}$ \quad Ben Lengerich$^{1,3}$ \\
\\
$^{1}$University of Wisconsin--Madison, $^{2}$Microsoft Research, $^{3}$Intelligible AI \\
\\
\texttt{jjia39@wisc.edu, chansingh@microsoft.com,} \\
\texttt{richcaruana@gmail.com, lengerich@wisc.edu}
}
\input{preamble}

\begin{document}

\ifcolmsubmission
\linenumbers
\fi

\maketitle

\begin{abstract}
 Identifying meaningful feature interactions is a central challenge in building accurate and interpretable models for tabular data.
  Generalized additive models (GAMs) have shown great success at modeling tabular data, but often rely on heuristic procedures to select interactions, potentially missing higher-order or context-dependent effects.
  To meet this challenge, we propose \method, a method that leverages tabular foundation models and post-hoc distillation methods.
  Our key intuition is that tabular foundation models implicitly learn rich, adaptive feature dependencies through large-scale representation learning.
  Given a dataset, \method first fits a tabular foundation model to the dataset, and then applies a post-hoc interaction attribution method to extract salient feature interactions from it.
  We evaluate these interactions by then using them as terms in a GAM.
  Across tasks, we find that interactions identified by \method lead to consistent improvements in downstream GAMs' predictive performance. Our results suggest that tabular foundation models can serve as effective, data-driven guides for interaction discovery, bridging high-capacity models and interpretable additive frameworks.  \footnote{Code to use \method and to reproduce our experiments is available on Github at \href{https://github.com/Clouddelta/tab-distill}{github.com/Clouddelta/tab-distill}.}

\end{abstract}

\section{Introduction}

As tabular machine learning becomes more widespread,
interpretable models are increasingly essential in high-stakes applications and for scientific discovery, where understanding how inputs jointly influence predictions is often as important as predictive accuracy~\citep{rudin2019stop,molnar2019interpretable,murdoch2019definitions}.
Generalized additive models (GAMs) are a widely used class of interpretable models that decompose predictions into feature-level effects, offering transparency and strong performance on tabular data~\citep{hastie1987generalized,lou2012intelligible,agarwal2021neural}.

While GAMs have shown strong performance,
they often struggle in real-world problems that involve feature interactions which cannot be captured solely through an additive model.
% by purely additive structure, and identifying a small, meaningful set of interactions remains a central challenge.
Existing approaches for selecting interactions to use GAMs typically rely on greedy heuristics or model-specific criteria, such as sequentially adding pairwise terms based on residual improvements.
These methods are limited in their ability to detect a small, meaningful set of interactions across diverse datasets.

Meanwhile, modern tabular foundation models (TFMs) have demonstrated an impressive capability to model rich feature interactions,
achieving it through large-scale pretraining and in-context learning~\citep{hollmann2022tabpfn,hollmann2025accurate,qu2025tabicl}.
While TFMs capture complex interactions, their large size and opaque structure make them difficult to interpret or directly deploy in settings that require interpretability.

\begin{figure*}[t]
\centering
\includegraphics[width=\textwidth,trim=0cm 6cm 0cm 0cm, clip]{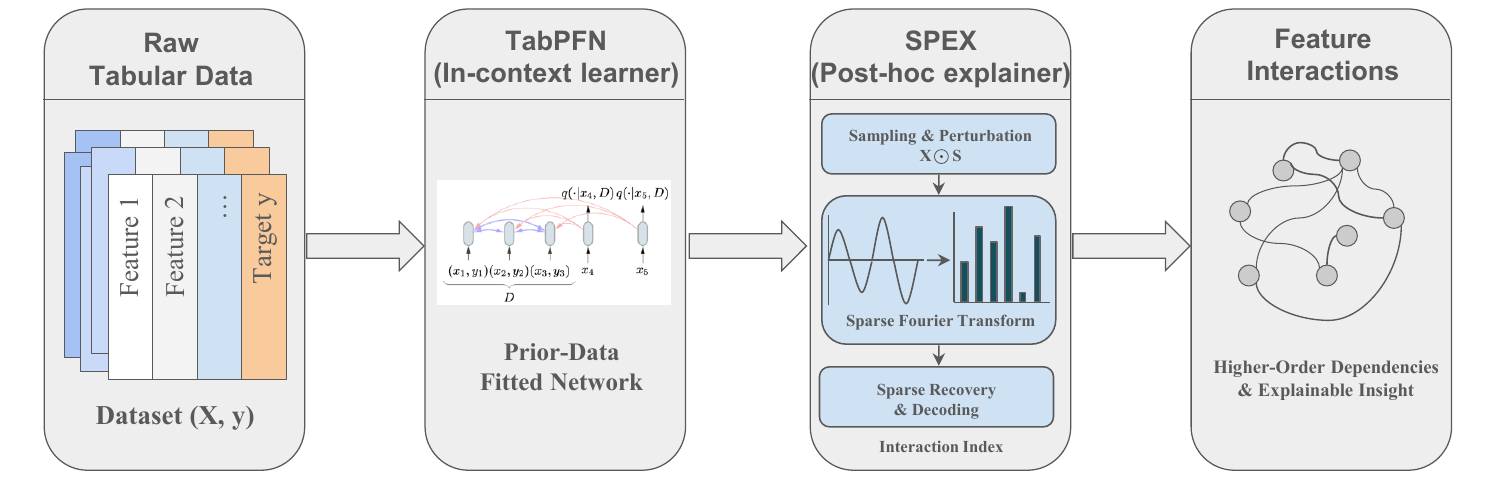}
\caption{\method utilizes the adaptive feature dependency structure learned from TFM. The distilled model is then passed through the post-hoc explainer SPEX for interaction feature discovery.}
\label{fig:intro}
\end{figure*}

To meet this challenge, we propose \method, a method that bridges TFMs and interpretable models by using TFMs to identify interactions that are then used as additive terms in an interpretable GAM (see Figure~\ref{fig:intro}).
% The key idea is to extract a ranked set of salient interactions from a fitted TFM via post-hoc attribution. 
Given a dataset, \method first fits a TFM.
Then, a scalable post-hoc interaction attribution method is applied to the fitted TFM to identify important feature interactions.
Finally, these interactions are incorporated as additive terms in a GAM.

% We conduct experiments on \method using a variety of TFMs (particularly TabPFN~\citep{hollmann2025accurate}) and post-hoc attribution methods (particularly the Faith-Banzhaf Interaction Index~\citep{tsai2023faith} computed via SPEX~\citep{kangspex2025}).
We conduct experiments on \method using TabPFN and post-hoc attribution methods (particularly the Faith-Banzhaf Interaction Index~\citep{tsai2023faith} computed via SPEX~\citep{kangspex2025}).
Across a range of real-world and synthetic benchmarks, we show that interactions selected by \method consistently improve the predictive performance of downstream GAMs compared to standard heuristic and post-hoc baselines. These results suggest that TFMs can serve as effective, data-driven guides for interaction discovery, enabling interpretable additive models to benefit from the representational power of modern TFMs without sacrificing interpretability.

\section{Related works}

\paragraph{GAMs}
Generalized Additive Models (GAMs)~\citep{hastie1987generalized} emerged as a generalization of Generalized Linear Models~\citep{nelder1972generalized} which include non-linear transformations of the input features. 
Traditional GAMs often use splines and backfitting~\citep{hastie1987generalized}, enhanced by penalized regression splines~\citep{wood2003thin} and fast fitting algorithms~\citep{wood2001mgcv}. Spline-based GAMs use the backfitting algorithm, iteratively updating each shape function to fit the residuals of others until convergence. More recent advances include Explainable Boosting Machines (EBMs)~\citep{lou2012intelligible, lou2013accurate, caruana2015intelligible}, which use decision trees to model shape functions via cyclic gradient boosting and are widely used to model treatment effects in medical data. On the other hand, Neural Additive Models \citep{agarwal2021neural} and related works~\citep{chang2021node, dubeyscalable, radenovicneural, xu2022sparse, enouensparse, Bouchiatetal24,yang2021gami} use multilayer perceptrons as non-linear transformations to model the shape functions.

\paragraph{Model-agnostic interaction detection}
We build off of the recently proposed SPEX~\citep{kangspex2025}.
We compare against a set of post-hoc methods based on game-theoretic notions, described in \cref{sec:results}~\citep{tsai2023faith}.
Other recent methods are model-specific, e.g. for trees~\citep{basu2018Iterative} or for a neural network~\citep{tsang2017detecting}.

\paragraph{Tabular foundation models (TFMs)}
TFMs adapt the in-context learning paradigm to predictive modeling for tabular data. Building on this idea, \citeauthor{hollmann2022tabpfn}~(\citeyear{hollmann2022tabpfn}) introduce TabPFN, a transformer-based model trained exclusively on synthetic tabular datasets that performs prediction via in-context learning, thereby eliminating task-specific optimization at test time. Subsequent work, TabPFN-2~\citep{hollmann2025accurate}, substantially improves scalability, supporting up to 10,000 samples and 500 features. More recently, TabPFN-2.5~\citep{grinsztajn2025tabpfn} further extends this regime to 50,000 samples and 2,000 features, while significantly accelerating inference through a proprietary multi-layer perceptron–based distillation engine.
Besides this line of work, a variety of models with different inference-time tradeoffs (e.g. TabICL~\citep{qu2025tabicl}) or different task settings, e.g. causal discovery~\citep{robertson2025pfn,bynum2025black,balazadeh2025causalpfn,ma2025foundation},
statistical inference~\citep{peyrard2025meta},
or for directly producing an interpretable model~\citep{mueller2024gamformer,zhuang2024learning}.

\paragraph{Distillation}

A variety of works have studied distilling pre-trained black-box models into interpretable models in various domains such as computer vision~\citep{ghosh2023distilling,ha2021adaptive},
speech~\citep{yang2023knowledge}, reinforcement learning~\citep{dispoto2025so}, natural-language processing~\citep{singh2023augmenting},
and tabular data~\citep{tan2018distill,si2024interpretabnet}.
In black-box models, researchers have found that distillation can improve the relative interpretability of the student model~\citep{han2023impact,liu2018improving}.

\section{Methods: \method}

\paragraph{Objective}
We seek to identify feature interactions from a supervised dataset that can improve performance when used in a GAM.
The structure of a GAM is given by: 
\begin{equation}\label{eq:gam-definition} 
%g\big(\mathbb{E}\big[Y | X\big]\big) = \beta + \sum\nolimits_{i=1}^p f_i(X_i) 
g ( \mathbb{E} [ y | x ] ) = \beta + \underbrace{\sum\nolimits_{j=1}^p f_j(x_j)}_{\text{univariate shape functions}} + \underbrace{\sum\nolimits_{I \in \mathcal I} f_I (x_I)}_{\text{interaction terms}}
\end{equation} 

where $x = (x_1, \dots, x_p) \in \inspace \subseteq \mathbb{R}^p$ is the input with $p$ features, $y\in \outspace \subseteq \mathbb{R}^m$ is the response variable, and $f_i: \mathbb{R} \to \mathbb{R}$ are univariate functions called \textit{shape functions} that capture the individual contributions of each feature. The intercept $\beta \in \mathbb{R}$ is a learnable bias term, and $g: \mathbb{R} \to \mathbb{R}$ is the link function that connects the expected outcome to the linear predictor, examples of which include the logit or softmax function for binary or multiclass classification or the identity function for linear regression. The shape functions $f_i$ allow for an interpretable representation of each feature's effect, akin to the role of coefficients in linear regression, thus enabling practitioners to inspect the learned non-linear relationships. 

The interaction terms $f_I$ map a set of two or more features $I$ to a continuous value.
Each of these interaction terms satisfies the properties of non-additive interactions~\citep{tsang2017detecting}, i.e., 
% Definition 1 (Non-additive Interaction). Consider a function $f(\cdot)$ with input variables $x_i, i \in[p]$, and an interaction $\mathcal{I} \subseteq[p]$. Then $\mathcal{I}$ is a non-additive interaction of function $f(\cdot)$ if and only if
there does not exist a set of functions $f_i(\cdot)$, for any $i \in I$, where $f_i(\cdot)$ does not depend on $x_i$, such that
$$
f(\mathbf{x})=\sum_{i \in I} f_i\left(\mathbf{x}_{[p] \backslash\{i\}}\right) .
$$
These interactions can improve predictive performance, but often add difficulty to interpretation, creating an incentive to minimize the number of interaction terms included.

We want to find the best GAM model with up to $K$-order interactions that minimizes the expected loss. We use the definition of interactions in \cite{lou2013accurate} to define our objective. For a dataset $\mathcal{D} = \{(x_i, y_i)\}_{i=1}^N$ of size $N$, each input
$x_i = (x_{i1}, \ldots, x_{ip}) \in \mathbb{R}^p$ consists of $p$ features and
$y_i$ denotes the corresponding response. We define the index set of
interactions up to order-$K$ as $\mathcal{U}_K = \bigcup_{k=1}^K \mathcal{U}^k$, where
$\mathcal{U}^k = \bigl\{ \{i_1, \ldots, i_k\} \,\big|\, 1 \le i_1 < \cdots < i_k \le p \bigr\}$ is the order-$k$ interaction index set.

Let $\mathcal{H}_K = \bigoplus_{u \in \mathcal{U}_K} \mathcal{H}_u$
denote the Hilbert space of functions of
$x \in \mathbb{R}^p$ with additive form 
\[
f(x) = \sum_{u \in \mathcal{U}_K} f_u(x_u)
\]
on up to K-dimensional shape functions. Our objective is to solve:
\[
\min_{f \in \mathcal{H}_K} \; \mathbb{E}[L(y, f(x))],
\]
where $L(\cdot,\cdot)$ is a non-negative convex loss function.

\paragraph{Method}
To identify such interactions $I$, we propose \method, which takes in a supervised dataset and outputs a ranked list of important feature interactions for the use of the downstream GAM.
\method leverages the intuition that tabular foundation models implicitly learn rich, adaptive feature dependencies through large-scale representation learning.

As its first step, \method fits a TFM to the dataset. Then, \method applies a post-hoc interaction attribution method to extract salient feature interactions from the fitted model.
While many choices can be made for the post-hoc attribution method, we
desire a method that does not require internal access to the TFM (so it is compatible with a variety of architectures, including those not yet released), is fairly sample-efficient (to avoid making too many computationally intensive calls to the TFM), and can effectively find predictive interactions. 
To meet these criteria, we 
use the Faith-Banzhaf Interaction Index (FBII) ~\citep{tsai2023faith}, calculated via SPEX~\citep{kangspex2025}. SPEX learns an approximate sparse Fourier transform surrogate $\hat{f}(x)\approx f(x)= \sum_{u \in \mathcal{U}_K} f_u(x_u)$. It efficiently finds a small set of $\mathbf{k}$ with $|\mathbf{k}| \ll n$,
denoted $\mathcal{K}$, and $\hat{F}(\mathbf{k})$ for each
$\mathbf{k} \in \mathcal{K}$ such that
\begin{equation*}
  \hat{f}(\mathbf{m}) = \sum_{\mathbf{k} \in \mathcal{K}}
(-1)^{\langle \mathbf{m}, \mathbf{k} \rangle} \hat{F}(\mathbf{k}),  
\end{equation*}
where $\mathbf{m}\in \mathbb{F}_2^n$ is a binary masking vector.

SPEX recovers sparse, low-order Fourier coefficients corresponding to feature interactions, from which we compute the
% standard interaction indices such as the
FBII as post-processing. By combining sparse Fourier analysis with structured masking and efficient decoding, SPEX avoids enumerating the exponential interaction space, enabling faithful interaction-based explanations in long-context settings beyond marginal effects.

With $\hat{f}$ and $\hat{F}$ identified by SPEX, \method then identifies the feature importance coefficients in the surrogate function and can specify the important interaction effects in the distilled foundation model.
\cref{alg:tabpfn_spex} summarizes the interaction search procedure of TabDistill.

\begin{algorithm}[tb]
  \caption{Interaction Selection via TabDistill}
  \label{alg:tabpfn_spex}
  \begin{algorithmic}
    \STATE {\bfseries Input:} Dataset $(X, y)$, number of interactions $k$
    \STATE {\bfseries Output:} Top-$k$ interactions $\mathcal{I}_{\text{top}}$ for a GAM model

    \STATE Fit TabPFN on $(X, y)$ to obtain model $\mathcal{A}$

    \STATE Initialize an empty multiset $\mathcal{C}$

    \FOR{each sample $(X_i, y_i)$ in $(X, y)$}
        \STATE $\mathcal{I}_i \leftarrow \text{SPEX}(\mathcal{A}, X_i, y_i)$
        \STATE Add all interactions in $\mathcal{I}_i$ to $\mathcal{C}$
    \ENDFOR

    \STATE Count the frequency of each interaction in $\mathcal{C}$ and select the $k$ most frequent interactions as $\mathcal{I}_{\text{top}}$ for downstream GAM fitting

    \STATE \textbf{return}\ $\mathcal{I}_{\text{top}}$
  \end{algorithmic}
\end{algorithm}

\section{Results}
\label{sec:results}

\subsection{Experimental setup}
\paragraph{Datasets}
We select datasets in TabArena~\citep{erickson2025tabarena}, TALENT~\citep{ye2024closer}, and PMLB~\citep{Olson2017PMLB} that have fewer than 2000 samples and fewer than 10 features. After removing deprecated datasets in PMLB, we included 35 regression task datasets and 44 classification task datasets.

\textbf{Baselines}
As a baseline, we run FAST~\citep{lou2013accurate}, the standard greedy interaction selection heuristic in the InterpretML package~\citep{lou2012intelligible,caruana2015intelligible,nori2019interpretml}.
As a second baseline, we compare against RuleFit~\citep{friedman2008predictive}, which selects rules from a random forest and ranks them based on their support (using the default hyperparameters implemented in the imodels package~\citep{singh2021imodels}).

For \method, we choose TabPFN-2 as the TFM to be distilled and SPEX with the feature interaction index FBII as the post-hoc explanation method. Since the computational cost of the post-hoc inference is high, we use a maximum interaction order of $k = 3$ , and a sample budget $B=500$ in SPEX. We also compare against different variations of post-hoc feature interaction detection methods, including the Faith-Shapley Interaction Index (FSII)~\citep{tsai2023faith},
the Shapley Taylor Interaction Index (STII)~\citep{sundararajan2020shapley},
the Banzhaf Interaction Index (BII)~\citep{grabisch2000equivalent},
the Shapley Interaction Index (SII)~\citep{grabisch2000equivalent},
Möbius Interactions (Mobius)~\citep{harsanyi1982simplified},
and Fourier interactions (Fourier)~\citep{ahmed1976orthogonal}. Unless otherwise specified, \method refers to \method with the FBII index.

\paragraph{Evaluation}
Evaluation is done by passing the resulting interactions to EBM~\citep{lou2012intelligible}, run through the InterpretML package~\citep{nori2019interpretml} using \texttt{outer\_bags} $= 4$ and \texttt{max\_bins} $= 256$; all other hyperparameters are set to their default values. Since all the methods allow an input of the number of interactions $N_{int}$ included in the model, we compare TabDistill with competing methods across three regression metrics: MSE, MAE, and $R^2$, and three classification metrics: Accuracy, F1 score and AUROC, with $N_{int}$ varying from 1 to 8.
For each metric and number of interactions, we rank nine methods, comprising two baselines (FAST and RuleFit) and seven TabDistill variants with different post-hoc feature interaction detection methods.

\subsection{Main result: \method identifies interactions that improve prediction performance}
\label{sec:main_result}

\begin{figure}[t]
\centering

\begin{subfigure}[t]{0.48\linewidth}
    \centering
    \includegraphics[width=\linewidth,trim=0cm 0cm 0cm 2.5cm,clip]{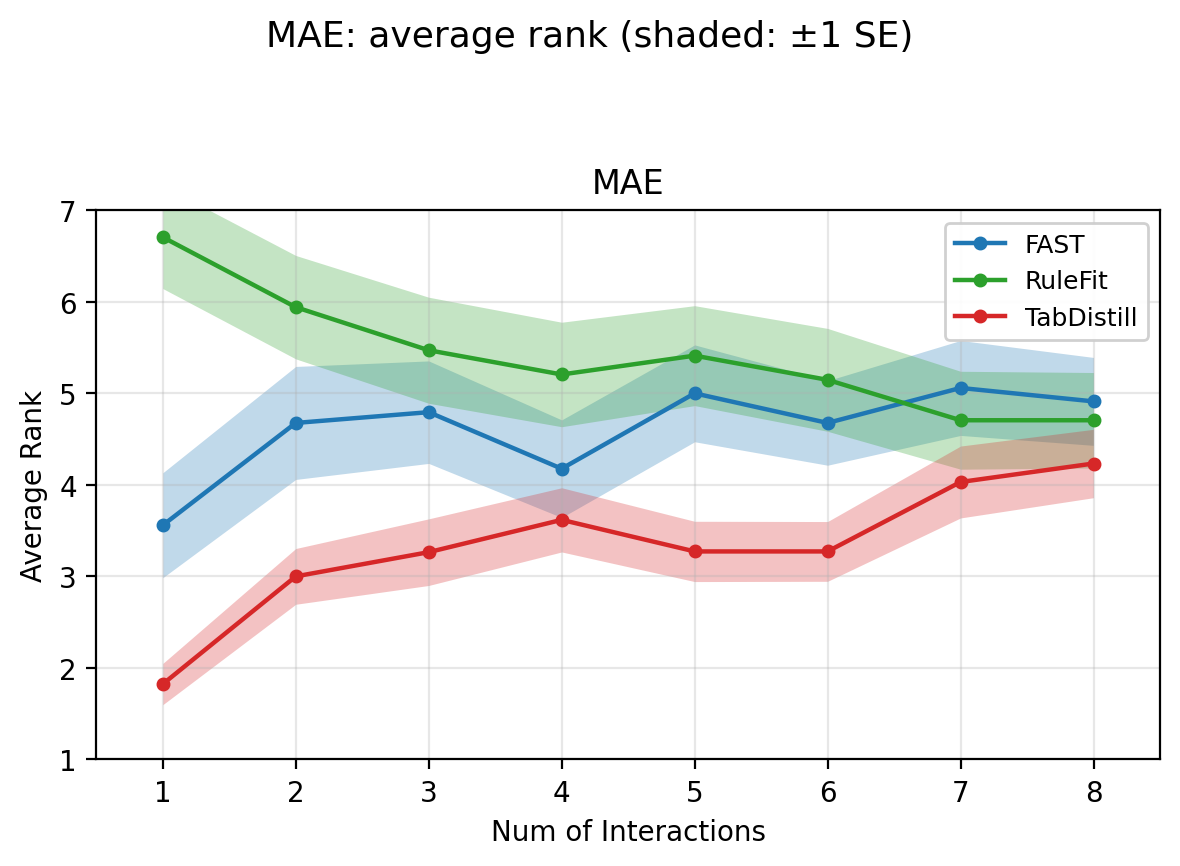}
    \caption{Regression tasks. Performance is evaluated by the average rank across datasets using the MAE metric. Lower is better.}
    \label{fig:rank_reg_mae}
\end{subfigure}\hfill
\begin{subfigure}[t]{0.48\linewidth}
    \centering
    \includegraphics[width=\linewidth,trim=0cm 0cm 0cm 2.5cm,clip]{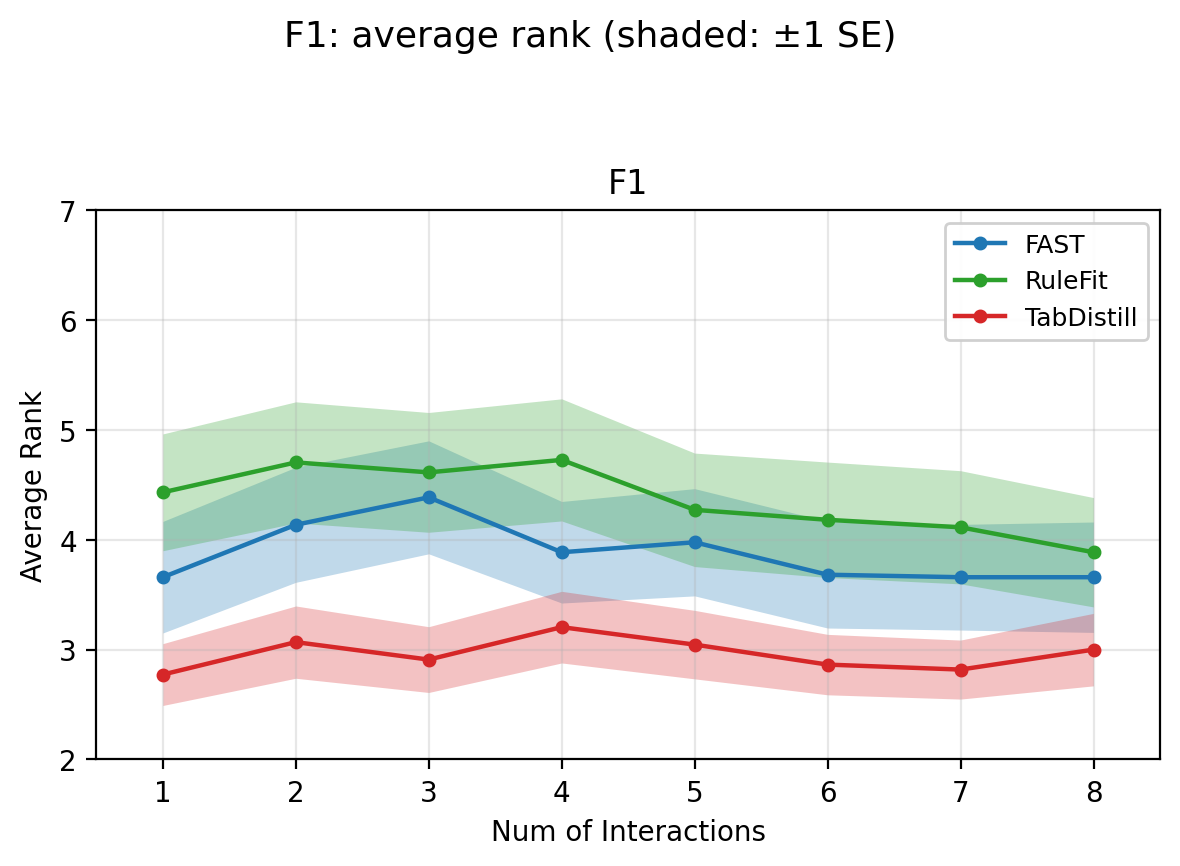}
    \caption{Classification tasks. Performance is evaluated by the average rank across datasets using the F1 score metric. Lower is better.}
    \label{fig:rank_cls_f1}
\end{subfigure}

\caption{\textit{Comparison of TabDistill against baseline methods.}
When the number of interactions is small, selecting informative interactions is critical. TabDistill remains superior across settings. As more interactions are included, sensitivity to the specific interaction set decreases.}
\label{fig:rank_comparison}
\end{figure}

\paragraph{Comparison to interaction selection baselines.}
Figure~\ref{fig:rank_comparison} reports the average ranks across all datasets, where \method achieves better performance than the baseline methods. A more detailed comparison is included in Appendix~\ref{sec:add_main_result}. Beyond predictive performance, \method also offers several advantages over FAST and RuleFit in terms of interaction modeling.

While FAST is computationally efficient for interaction selection, it is restricted to identifying pairwise interactions. In contrast, \method is able to discover higher-order interactions, enabling it to capture more complex dependency structures that are not accessible to FAST.

% \paragraph{Comparison to RuleFit.}
As a tree-based approach, RuleFit can implicitly capture higher-order interactions. However, our results show that the interactions identified by RuleFit do not translate into comparable predictive improvements. In contrast, \method selects higher-order interactions that more effectively improve downstream predictive performance. 

\paragraph{Comparison across interpretable and black-box models.}
Table~\ref{tab:avg_rank_models} summarizes the average rank of different models across PMLB regression datasets. Within the interpretable models (top section), EBM consistently achieves the best performance and even outperforms the black-box model XGBoost in this dataset collection. We also point out that PyGAM~\citep{pygam}, despite using the same \method-selected interactions, performs worse than EBM across all metrics. Overall, \method helps reduce the performance gap between interpretable and black-box models (bottom section).

\subsection{\method Ablations}
\paragraph{FBII outperforms other indices for distillation.}
Continuing the analysis from \cref{sec:main_result}, the FBII index achieves a low average rank while also exhibiting a small variance across all post-hoc interaction indices (\cref{tab:rank_mean_variance}), indicating both strong predictive performance and high stability across different interaction budgets. These results suggest that FBII provides a particularly reliable choice of index for interaction selection in \method. Consequently, unless otherwise specified, we adopt the FBII index in subsequent experiments when comparing \method against baseline methods. 

\begin{table}[t]
\centering
\small
\begin{tabular}{lccc}
\toprule
Model & MAE & MSE & $R^2$ \\
\midrule
\textbf{EBM (with \method)} & 2.26 & 2.37 & 2.37  \\
PyGAM (with \method)        & 4.65 & 4.69 & 4.69  \\
Linear Regression           & 4.33 & 4.19 & 4.19  \\
Decision Tree               & 4.78 & 4.78 & 4.78  \\
\midrule
XGBoost                     & 3.30 & 3.30 & 3.30  \\
TabPFN                      & 1.63 & 1.63 & 1.63  \\
\bottomrule
\end{tabular}
\caption{
\textit{Average rank of models across PMLB regression datasets (lower is better).}
EBM and PyGAM are implemented with 4 interactions selected from \method.
Compared to other interpretable models (top section), EBM achieves superior performance,
helping to close the gap between interpretable and black-box models (bottom section).
}
\label{tab:avg_rank_models}
\end{table}

\paragraph{TabDistill works well with a limited SPEX query budget.}
We evaluate the effect of the sample size used for interaction selection on TabArena regression datasets. For each method, we vary the number of samples used to identify interactions from 100 to 500, and compare the resulting selected interaction set against the reference set obtained by the same method using 500 samples. Table~\ref{tab:summary_by_index} shows that the interaction sets identified by \method under smaller query budgets remain close to those obtained with 500 query samples. Consistent with this, varying the number of SPEX query samples from 100 to 500 leads to only minor changes in downstream predictive performance, with average rank variations typically within about 0.5 across interaction indices. The corresponding performance comparison is shown in Appendix Figure~\ref{fig:sample_size}.

% \begin{figure}[ht]
%   \vskip 0.2in
%   \begin{center}
%     \centerline{\includegraphics[width=\columnwidth]{figs/sensitivity_sample_size.pdf}}
%     \caption{
%         \textit{Stability of interaction selection with respect to the interaction query budget of the explainer.}
%         We vary the number of interaction queries used for interaction detection in \method from 100 to 500 and evaluate downstream predictive performance on TabArena datasets, reported as average rank across datasets. Across all indices, rank variations remain within approximately 0.5.
%         \cs{Can you better motivate this plot in the writing / here? Why does this matter? }
%     }
%     \label{fig:sample_size}
%   \end{center}
% \end{figure}

\begin{figure}[t]
\centering

\begin{minipage}[t]{0.48\columnwidth}
    \centering
    \includegraphics[width=\linewidth,trim=0cm 0cm 0cm 0.6cm,clip]{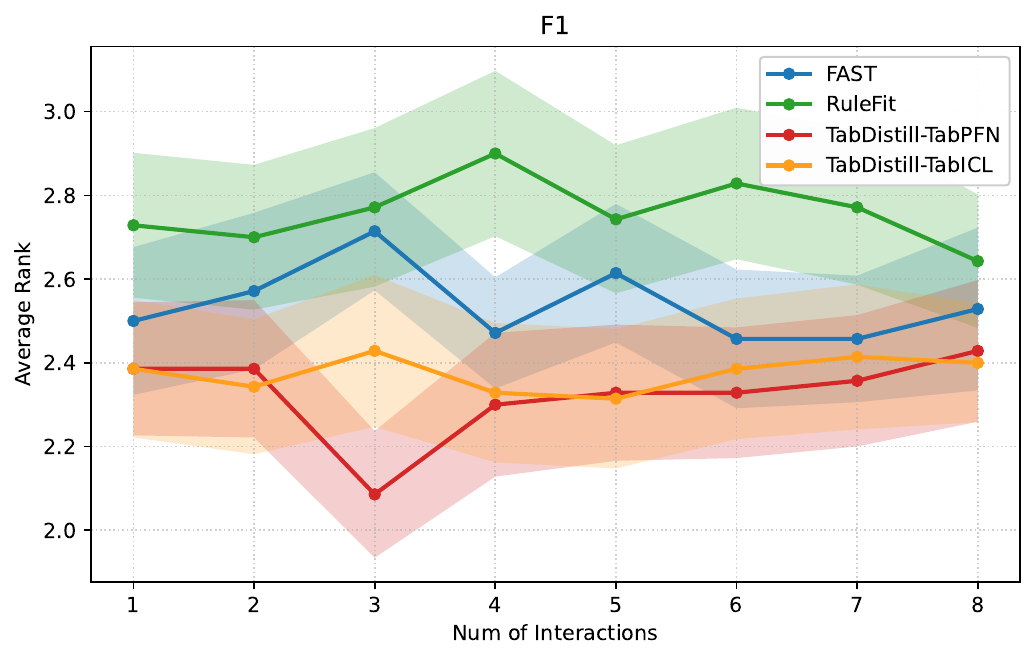}
\end{minipage}\hfill
\begin{minipage}[t]{0.48\columnwidth}
    \centering
    \includegraphics[width=\linewidth,trim=0cm 0cm 0cm 0.6cm,clip]{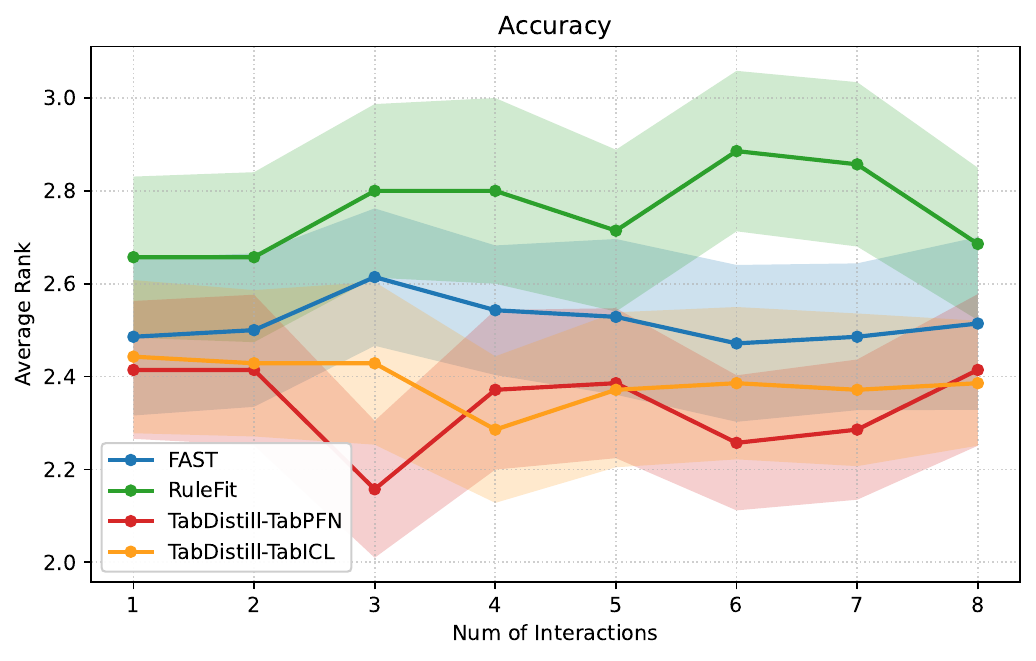}
\end{minipage}

\caption{\textit{Comparison of \method on TabICL and TabPFN against baseline methods.}
Left: performance evaluated by average rank across datasets using the F1 score metric. Right: performance evaluated by average rank across datasets using the accuracy metric.}
\label{fig:TabICL}
\end{figure}

\paragraph{\method works with different TFMs.}
In order to evaluate the generality of \method,
we also implemented TabDistill on another TFM, TabICL~\citep{qu2025tabicl}, and evaluated it on 42 classification datasets from PMLB.
Figure~\ref{fig:TabICL} shows that TabDistill remains effective beyond the original backbone: across different interaction budgets, TabDistill consistently achieves the best or near-best average rank compared with FAST and RuleFit. These results suggest that the benefit of TabDistill is not tied to a specific TFM and transfers well to other TFMs.

\begin{table}[t]
\centering
\small

\label{tab:combined_stats}
\begin{tabular}{l c c c c c}
\toprule
\multirow{2}{*}{Method} & \multicolumn{2}{c}{MAE} & & \multicolumn{2}{c}{$F1$} \\
\cmidrule{2-3} \cmidrule{5-6}
& Mean & Variance & & Mean & Variance \\
\midrule
TabDistill    & \textbf{3.52} & 0.60 & & 3.05 & 0.07 \\
FAST    & 4.64 & 0.20 & & 3.81 & 0.08 \\
RuleFit & 5.16 & 0.63 & & 4.31 & 0.08 \\
\midrule
BII     & 4.28 & 1.57 & & 3.05 & 0.16 \\
Fourier & 4.61 & 1.88 & & 3.41 & 0.07 \\
FSII    & 3.65 & 0.75 & & 3.14 & 0.03 \\
Mobius  & 3.71 & 0.93 & & 3.06 & 0.06 \\
SII     & 4.08 & 1.19 & & 3.12 & 0.11 \\
STII    & 3.64 & 0.79 & & \textbf{2.96} & 0.03 \\
\bottomrule
\end{tabular}
\caption{\textit{Performance summary for GAMs with different interaction selection methods.} We aggregate the ranks of all methods across datasets and interaction budgets $N_{int}$. TabDistill with FBII index (first line) shows a strong overall performance across tasks for interaction selection.}
\label{tab:rank_mean_variance}
\end{table}

\begin{table}[H]
\centering
\small
\begin{tabular}{lccccc}
\toprule
Method & \multicolumn{5}{c}{Sample size for interaction selection} \\
 & 100 & 200 & 300 & 400 & 500 \\
\midrule
\method  & 0.60 & 0.82 & 0.88 & 0.75 & 1.00 \\
FAST     & 0.58 & 0.55 & 0.59 & 0.60 & 1.00 \\
RuleFit  & 0.38 & 0.50 & 0.54 & 0.51 & 1.00 \\
\midrule
BII      & 0.55 & 0.77 & 0.93 & 0.85 & 1.00 \\
Fourier  & 0.70 & 0.88 & 0.89 & 0.90 & 1.00 \\
FSII     & 0.65 & 0.90 & 0.90 & 0.88 & 1.00 \\
Mobius   & 0.62 & 0.82 & 0.93 & 0.85 & 1.00 \\
SII      & 0.53 & 0.72 & 0.80 & 0.80 & 1.00 \\
STII     & 0.62 & 0.77 & 0.90 & 0.80 & 1.00 \\
\bottomrule
\end{tabular}
\caption{\textit{Stability of interaction selection across sample sizes.}
Each entry reports the proportion of the 8 selected interactions that overlap with the reference set obtained by the same method using 500 samples. Higher is better.}
\label{tab:summary_by_index}
\end{table}

\subsection{Analysis: TabPFN effectively captures interaction structure in synthetic domains}

To understand why \method is effective, we ask a more basic question: can TabPFN reliably recover non-additive interaction structure from tabular data? We study this on two families of controlled synthetic problems. First, we consider Fourier-sparse functions, which directly match the structural assumption exploited by SPEX. Second, we consider tree-structured teacher rules, which provide a complementary class of non-smooth interaction patterns. Across both settings, TabPFN consistently captures the underlying interaction structure more accurately compared to baseline methods, supporting its role as a strong teacher model for \method.

\subsubsection{Scenario A: Fourier-sparse Structured Interaction}
\begin{figure*}[t]
\centering
\includegraphics[width=\textwidth]{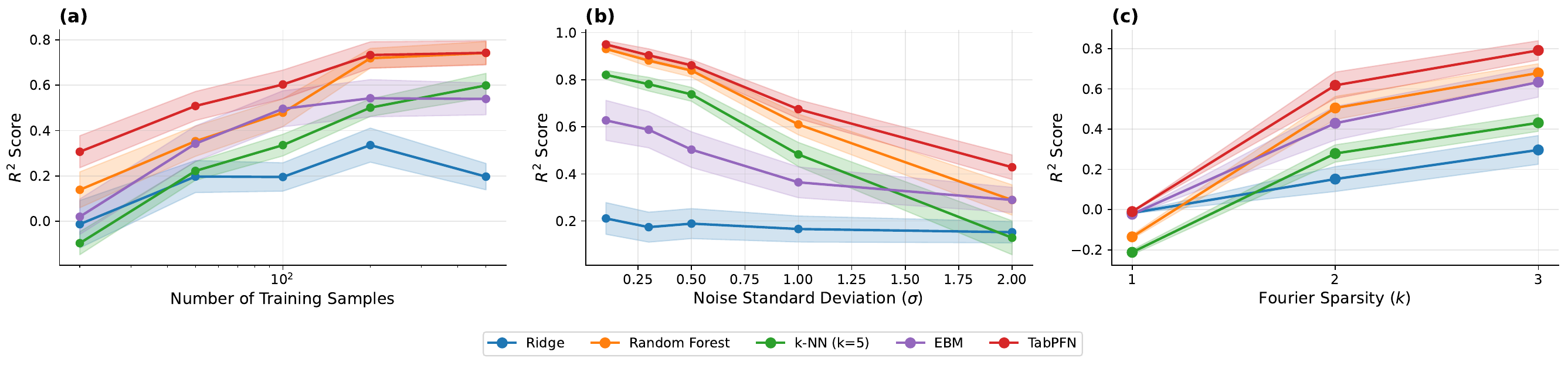}
\caption{\textit{Performance comparison of TabPFN against baseline methods on Fourier-sparse functions.}
\textbf{(a) Low-data regime;} \textbf{(b) Noise robustness;}  \textbf{(c) Extreme sparsity.}
}
\label{fig:fourier_sparse_results}
\end{figure*}

Our first set of experiments focuses on Fourier-sparse functions, since SPEX is motivated by the observation that explanation value functions are often sparse in the Boolean Fourier basis. We generate binary inputs and construct targets using a sparse Fourier expansion with low-order subsets, then evaluate TabPFN under low-data, noisy, and extreme-sparsity regimes. These experiments test whether TabPFN can recover a small number of relevant interactions from limited samples in a large combinatorial space. As shown in Figure~\ref{fig:fourier_sparse_results}, TabPFN performs strongly across all three regimes, especially in the extreme-sparsity setting where only a few active terms must be identified among exponentially many candidates.
% each averaged over 10 random seeds with standard error reported.

% \caption{\textit{Performance comparison of TabPFN against baseline methods on Fourier-sparse functions.}
% \textbf{(a) Low-data regime:} $R^2$ score vs.\ training set size with $n=10$ features and $k=3$ Fourier sparsity. TabPFN demonstrates superior sample efficiency, achieving high performance with as few as 20 training samples. \textbf{(b) Noise robustness:} $R^2$ score vs.\ noise standard deviation $\sigma$ with $n=8$ features, $k=3$ sparsity, and 300 training samples. TabPFN maintains robust performance across different signal-to-noise ratios. \textbf{(c) Extreme sparsity:} $R^2$ score vs.\ Fourier sparsity $k$ with $n=15$ features and 400 training samples. With only $k \in \{1,2,3\}$ non-zero coefficients out of $2^{15}=32{,}768$ possible Fourier terms, TabPFN successfully identifies the minimal feature interactions while baseline methods struggle with this ``needle in haystack'' setting.
% All results are averaged over 30 random seeds with shaded regions showing standard error. 
% }

\textbf{Dataset Generation:} Binary input features $\mathbf{x} \in \{0,1\}^n$ were sampled uniformly at random (i.e., each feature was an independent Bernoulli$(0.5)$ random variable). Target functions were generated using the Walsh-Hadamard (Fourier) basis. A function is $k$-sparse if it can be expressed as 
\begin{equation}
y = \sum_{i=1}^k c_i \chi_{S_i}(\mathbf{x}) + \varepsilon,
\end{equation}
where $\chi_S(\mathbf{x}) = (-1)^{\sum_{j \in S} x_j}$ are basis functions indexed by feature subsets $S_i$, coefficients $c_i \sim \mathcal{N}(0, 2^2)$, and $\varepsilon \sim \mathcal{N}(0, \sigma^2)$ is Gaussian noise. For each function, we randomly selected $k$ subsets $S_i$ (always including the empty set for the constant term) with interaction order $|S_i| \leq 3$.

\textbf{Experiment 1 -- Low-Data Regime:} With $n=10$ features, $k=3$ Fourier sparsity, and $\sigma=0.5$, we varied training set size from 20 to 500 samples while holding the test set at 200 samples. This tests few-shot learning capability.

\textbf{Experiment 2 -- Noise Robustness:} With $n=8$ features, $k=3$ sparsity, 300 training and 200 test samples, we varied noise standard deviation $\sigma \in \{0.1, 0.3, 0.5, 1.0, 2.0\}$ to assess robustness to different signal-to-noise ratios.

\textbf{Experiment 3 -- Extreme Sparsity:} With $n=15$ features ($2^{15} = 32{,}768$ possible Fourier terms), we tested $k \in \{1, 2, 3\}$ with 400 training and 200 test samples and $\sigma=0.3$. This ``needle in a haystack'' setting evaluates the ability to identify minimal feature interactions in high dimensions.

\textbf{Baselines:} We compared TabPFN against standard ML models using their default hyperparameters:
Ridge regression (with 5-fold cross-validation to select the hyperparameter $\alpha \in \{0.01, 0.1, 1.0, 10.0, 100.0\}$ ($\alpha=1.0$), Random Forest (100 trees, max depth 10, adjusted for small sample sizes), $k$-Nearest Neighbors ($k=5$ with distance weighting, adjusted for small sample sizes), and EBM using default hyperparameters. Performance was measured using the $R^2$ score.

\subsubsection{Scenario B: Tree Structured Interaction}
\begin{figure*}[!t]
    \centering
    \includegraphics[width=\textwidth,trim=0 0.3cm 0 0.2cm,clip]{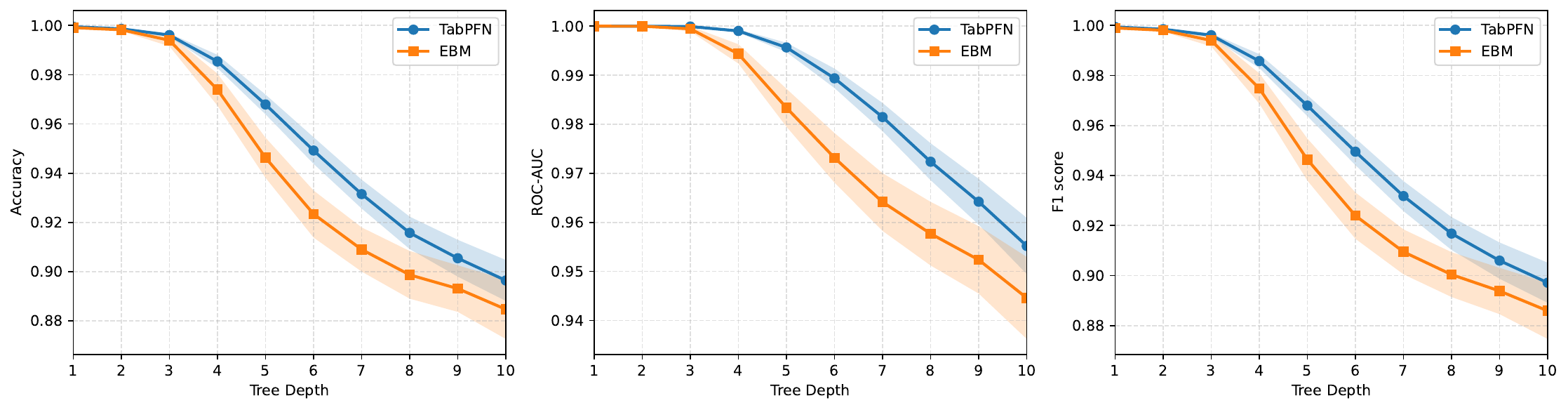}
    \caption{
        We simulate tree-structured data with $n=10{,}000$ samples and $p=15$ features, varying the decision tree depth from 1 to 10. As the decision boundary becomes increasingly complex and non-smooth, TabPFN more faithfully approximates data generated from tree-structured decision rules. All results are averaged over 20 random seeds.
    }
    \label{fig:tree}
    \vspace{-0.2cm}
\end{figure*}

We then consider tree-structured teacher rules. We train a decision tree on synthetic classification data, replace the original labels with the tree's predictions, and then compare how well TabPFN and EBM recover the teacher's decision function. By increasing tree depth, we progressively increase the complexity of the induced interaction structure. As shown in Figure~\ref{fig:tree}, TabPFN more faithfully matches the teacher across depth levels, indicating that its interaction-learning advantage extends beyond Fourier-sparse functions to non-smooth tree-based rules.

\textbf{Dataset Generation:} We generated synthetic binary classification data using \texttt{make\_classification} with $n=10000$ samples and $p=15$ features, of which $p_{\mathrm{inf}}=10$ were informative. The data were randomly split into training and test sets with a 7:3 ratio. To construct tree-structured target functions, we first trained a decision tree classifier on the training set using the original labels, and then replaced the original labels with the tree's predictions. This yields pseudo-labels $\tilde{y} = f_{\mathrm{tree}}(X)$, where $f_{\mathrm{tree}}$ denotes the fitted decision tree. The student models are evaluated on how well they recover the teacher's decision rule, rather than the original data-generating labels.
% \begin{figure*}[ht]
%   \vskip 0.2in
%   \begin{center}
%     \centerline{\includegraphics[width=\textwidth]{figs/tabpfn_vs_ebm_depth_metrics_ci.pdf}}
%     \caption{
%         We simulate tree-structured data with $n=10{,}000$ samples and $p=15$ features, varying the decision tree depth from 1 to 10. As the decision boundary becomes increasingly complex and non-smooth, TabPFN more faithfully approximates data generated from tree-structured decision rules. All results are averaged over 20 random seeds with shaded regions showing standard error.
%     }
%     \label{fig:tree}
%   \end{center}
% \end{figure*}

\textbf{Experiment -- Tree-Structured Rules:} We varied the maximum depth of the teacher tree over $D \in \{1,2,\dots,10\}$. Increasing $D$ produces progressively more complex and non-smooth decision boundaries, corresponding to higher-order and more localized interaction structures. For each depth, we trained student models on training set with pseudo-labels $(X_{\mathrm{train}}, \tilde{y}_{\mathrm{train}})$ and evaluated them on the testing set with pseudo-labels$(X_{\mathrm{test}}, \tilde{y}_{\mathrm{test}})$.

\textbf{Baselines:} We compared TabPFN against EBM as a student model for recovering the teacher rule. EBM was trained with budget set \texttt{num\_interactions} to $d_{\mathrm{inf}}$, using \texttt{outer\_bags}$=4$ and \texttt{max\_bins}$=256$. We ran the FAST algorithm for interaction detection in EBM. Performance was measured using classification accuracy, AUROC, and F1 score.

\subsection{Case study: Predicting used Fiat 500 car price}
We conduct a regression task on \textit{Another Dataset on used Fiat 500 (1538 rows)}, a used-car price prediction dataset from the TabArena benchmark (OpenML task ID 363615). The dataset contains 1,538 samples, each corresponding to an individual vehicle listing collected from a specialized second-hand car marketplace. It includes seven predictor variables, comprising both continuous and discrete features that describe vehicle characteristics and seller-related information. The prediction target is the vehicle's selling price.

We compare both predictive performance and interpretability across FAST, RuleFit, and TabDistill. For the experimental setup, we include the top 10 interactions identified by each method and train an EBM using these interactions. We use the official training/testing split provided by the OpenML task with a train--test ratio of 2:1, and evaluate performance on the test set. The EBM is fitted with \texttt{outer\_bags}=4 and \texttt{max\_bins}=256, while all other hyperparameters are set to their default values. Among all methods, TabDistill achieves the best overall performance, with an MAE of 536.1 and \(R^2\) score of 0.86.

% In terms of interpretability, TabDistill successfully identifies three-way or higher-order interactions involving seller location information. For example, TabDistill discovers an interaction among longitude, latitude, and model type, i.e., the \texttt{(longitude, latitude, model\_type)} interaction. The corresponding visualization is shown in \cref{fig:case_study}. We observe that the distribution of car prices differs depending on whether the vehicle is a sport model, and for sport models, prices exhibit spatial heterogeneity across regions. In contrast, FAST is limited to capturing only pairwise interactions, making it infeasible to model the joint effect of full location information with other features. RuleFit also fails to recover this interaction. The full performance table and the full list of the top 8 interactions identified by each method are provided in Appendix \ref{sec:case_study}.
In terms of interpretability, TabDistill identifies a higher-order interaction among longitude, latitude, and model type, i.e., \texttt{(longitude, latitude, model\_type)}. The selling price patterns in sport and non-sport models are different in the geographic space that is partitioned into \(2\times2\) cells using the median longitude and latitude. In contrast, FAST is limited to pairwise interactions and therefore cannot directly capture the joint effect of two-dimensional location and model type, and RuleFit fails to recover this interaction. The full performance table and the top 10 interactions identified by each method are provided in Appendix \ref{sec:case_study}.

\FloatBarrier
\section{Discussion}

Overall, \method provides a simple and modular way to transfer interaction structure from high-capacity tabular models to interpretable additive models, offering a practical path toward models that are both accurate and interpretable. Its utility may also extend beyond GAMs to settings where interactions themselves are scientifically meaningful~\citep{basu2018Iterative}.
Additionally, these interactions may be incorporated using domain knowledge in a human-in-the-loop workflow, e.g.~\citep{wang2021gam,feng2026human}.

While effective, \method inherits biases and failure modes of the underlying TFM, so spurious or unstable interactions may be transferred, and interaction attribution is often computationally expensive on large datasets. Future work could improve the performance--efficiency tradeoff, for example by using an intermediate distillation model as in ProxySPEX~\citep{butler2025proxyspex}, or by training TFMs to directly output important interactions rather than relying on post-hoc distillation.

% Overall, \method provides a simple and modular framework for bridging high-capacity tabular models and interpretable additive models, offering a practical path toward models that are both accurate and interpretable.
% Going forward, we expect \method to improve, both in performance and efficiency, as modern TFMs improve.
% Additionally, though we select interactions to be used in GAMs, the interactions could be useful in other contexts where interactions themselves are important, e.g., in scientific discovery~\citep{basu2018Iterative}.

% While effective, \method has some limitations.
% First, \method inherits biases and failure modes of the underlying TFM model: if the TFM model learns spurious or unstable interactions, these may be propagated to the GAM. Second, interaction attribution remains computationally expensive for large datasets.
% Nevertheless, in high-stakes settings where interpretability is necessary, this extra computational cost is often tolerable.

% Future work could explore better efficiency tradeoffs, e.g., using an intermediate model for distillation as in ProxySPEX~\citep{butler2025proxyspex}.
% Alternatively, TFMs could be trained to directly output important interactions, bypassing the need for post-hoc distillation altogether.

\newpage
% \section*{Acknowledgements}

% \section*{Impact Statement}

% This work introduces a method for extracting feature interactions from TFMs and using them in GAMs, with the goal of improving predictive performance while retaining interpretability.
% The approach inherits key limitations and risks from the underlying TFMs: if these models learn spurious, unstable, or biased interactions, such effects may be propagated and given unwarranted interpretive weight in the distilled GAM. In addition, interaction attribution requires repeated evaluations of a large model, which introduces non-trivial computational and energy costs, particularly if applied to larger datasets or higher-order interactions. These concerns highlight the need for careful validation, domain expertise, and judicious use of the method in sensitive or high-stakes settings.

\bibliography{main.bib}
\bibliographystyle{colm2026_conference}

\appendix

\input{appendix}

\end{document}

%% file: preamble.tex
\usepackage[ampersand]{easylist}
\ListProperties(Hide2=1,Hide3=2,Progressive*=.5cm,Numbers3=l, Numbers4=r,FinalMark3={)})
\usepackage{etoolbox}
\AtBeginEnvironment{easylist}{\ListProperties(Start1=1)}

\usepackage{xcolor}

\definecolor{cblue}{RGB}{8, 85, 153}

\usepackage{amsmath}
\usepackage[capitalise]{cleveref}
\usepackage{float}
\usepackage{placeins} %FloatBarrier
\usepackage{adjustbox}

\usepackage{makecell}
\usepackage[ampersand]{easylist}
\usepackage{todonotes}
\usepackage{caption}
\usepackage{xspace}
\usepackage{markdown}
\usepackage{multirow}
\usepackage{multicol}
\usepackage{graphicx}
\usepackage{subcaption}
\usepackage{titlesec}
\usepackage{tabularx}
\usepackage{array}

\titlespacing*{\paragraph}{0pt}{0.6em}{0.6em}

\usepackage{algorithmic}
\usepackage{algorithm}

\newcommand{\method}{TabDistill\xspace}

\usepackage{adjustbox}

\newcommand{\inspace}{\mathcal{X}}
\newcommand{\outspace}{\mathcal{Y}}

%% file: appendix.tex
\newpage
\appendix
\onecolumn
\counterwithin{figure}{section}
\counterwithin{table}{section}
\renewcommand{\thefigure}{A\arabic{figure}}
\renewcommand{\thetable}{A\arabic{table}}

\section{Appendix}
\FloatBarrier

This appendix provides additional benchmark results, ablation studies, and details for the Fiat 500 case study.

\subsection{Additional Results for Main Experiments}\label{sec:add_main_result}

We report the full rank breakdown for the main benchmark experiments. Across both regression and classification tasks, \method consistently outperforms FAST and RuleFit over a wide range of interaction budgets.
\begin{table}[!htbp]
\centering
\begin{minipage}[t]{0.48\textwidth}
\centering
\small
\renewcommand{\arraystretch}{1.15}
\begin{tabular}{ccccc}
\toprule
Metric & $N_{int}$ & \method & FAST & RuleFit \\
\midrule
\multirow{8}{*}{MAE}
 & 1 & \textbf{1.82} & 3.56 & 6.71 \\
 & 2 & \textbf{3.00} & 4.68 & 5.94 \\
 & 3 & \textbf{3.26} & 4.79 & 5.47 \\
 & 4 & \textbf{3.62} & 4.18 & 5.21 \\
 & 5 & \textbf{3.27} & 5.00 & 5.41 \\
 & 6 & \textbf{3.27} & 4.68 & 5.15 \\
 & 7 & \textbf{4.03} & 5.06 & 4.71 \\
 & 8 & \textbf{4.23} & 4.91 & 4.71 \\
\midrule
\multirow{8}{*}{MSE}
 & 1 & \textbf{1.88} & 3.26 & 6.50 \\
 & 2 & \textbf{3.35} & 4.41 & 5.47 \\
 & 3 & \textbf{3.41} & 4.71 & 5.65 \\
 & 4 & \textbf{3.82} & 4.47 & 4.71 \\
 & 5 & \textbf{3.88} & 4.47 & 4.50 \\
 & 6 & \textbf{3.76} & 4.18 & 4.85 \\
 & 7 & \textbf{3.88} & 4.41 & 4.47 \\
 & 8 & \textbf{4.40} & 4.62 & 4.62 \\
\midrule
\multirow{8}{*}{$R^2$}
 & 1 & \textbf{1.88} & 3.26 & 6.50 \\
 & 2 & \textbf{3.35} & 4.41 & 5.47 \\
 & 3 & \textbf{3.41} & 4.71 & 5.65 \\
 & 4 & \textbf{3.82} & 4.47 & 4.71 \\
 & 5 & \textbf{3.88} & 4.47 & 4.50 \\
 & 6 & \textbf{3.76} & 4.18 & 4.85 \\
 & 7 & \textbf{3.88} & 4.41 & 4.47 \\
 & 8 & \textbf{4.40} & 4.62 & 4.62 \\
\bottomrule
\end{tabular}
\caption{Regression tasks.}
\label{tab:rank_mae_mse_r2_fbii_fast_rulefit}
\end{minipage}
\hfill
\begin{minipage}[t]{0.48\textwidth}
\centering
\small
\renewcommand{\arraystretch}{1.15}
\begin{tabular}{ccccc}
\toprule
Metric & $N_{int}$ & \method & FAST & RuleFit \\
\midrule
\multirow{8}{*}{Accuracy}
 & 1 & \textbf{2.45} & 3.89 & 4.64 \\
 & 2 & \textbf{2.77} & 3.82 & 4.64 \\
 & 3 & \textbf{2.61} & 4.55 & 4.68 \\
 & 4 & \textbf{3.00} & 4.48 & 5.09 \\
 & 5 & \textbf{2.70} & 4.32 & 4.70 \\
 & 6 & \textbf{2.59} & 4.14 & 4.82 \\
 & 7 & \textbf{2.45} & 4.05 & 4.55 \\
 & 8 & \textbf{2.59} & 4.05 & 4.23 \\
\midrule
\multirow{8}{*}{F1 score}
 & 1 & \textbf{2.77} & 3.66 & 4.43 \\
 & 2 & \textbf{3.07} & 4.14 & 4.70 \\
 & 3 & \textbf{2.91} & 4.39 & 4.61 \\
 & 4 & \textbf{3.20} & 3.89 & 4.73 \\
 & 5 & \textbf{3.05} & 3.98 & 4.27 \\
 & 6 & \textbf{2.86} & 3.68 & 4.18 \\
 & 7 & \textbf{2.82} & 3.66 & 4.11 \\
 & 8 & \textbf{3.00} & 3.66 & 3.89 \\
\midrule
\multirow{8}{*}{AUROC}
 & 1 & \textbf{2.80} & 4.05 & 4.91 \\
 & 2 & \textbf{3.36} & 3.91 & 4.59 \\
 & 3 & \textbf{3.25} & 4.66 & 4.45 \\
 & 4 & \textbf{3.64} & 3.82 & 4.68 \\
 & 5 & 4.11 & \textbf{4.00} & 4.16 \\
 & 6 & 4.00 & \textbf{3.66} & 4.16 \\
 & 7 & \textbf{3.77} & 4.05 & 4.05 \\
 & 8 & \textbf{3.59} & 3.77 & 4.00 \\
\bottomrule
\end{tabular}
\caption{Classification tasks.}
\label{tab:rank_acc_f1_auroc_fbii_fast_rulefit}
\end{minipage}
\end{table}

\begin{figure*}[t]
\centering
\begin{subfigure}[t]{0.48\textwidth}
    \centering
    \includegraphics[width=\linewidth,trim=0cm 0cm 0cm 2.5cm,clip]{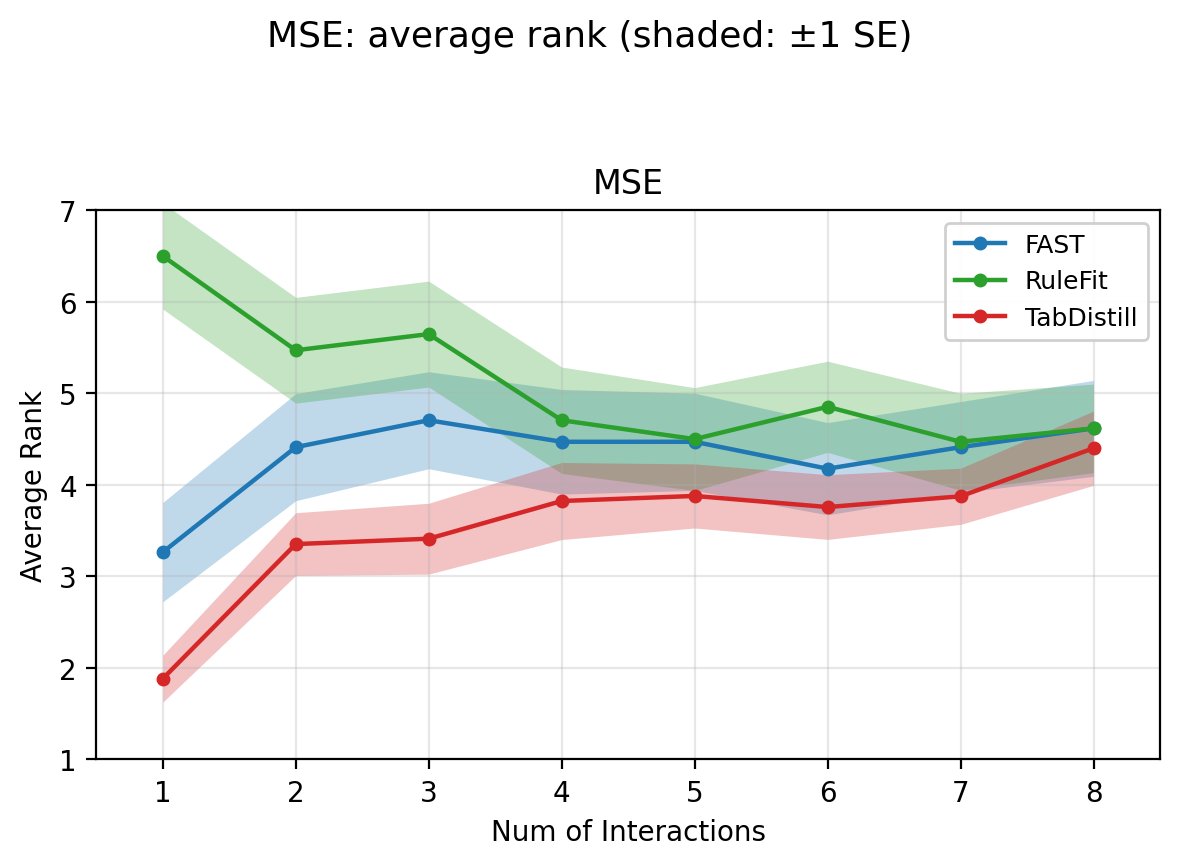}
    \caption{Regression tasks evaluated by MSE.}
    \label{fig:rank_reg_mse}
\end{subfigure}\hfill
\begin{subfigure}[t]{0.48\textwidth}
    \centering
    \includegraphics[width=\linewidth,trim=0cm 0cm 0cm 2.5cm,clip]{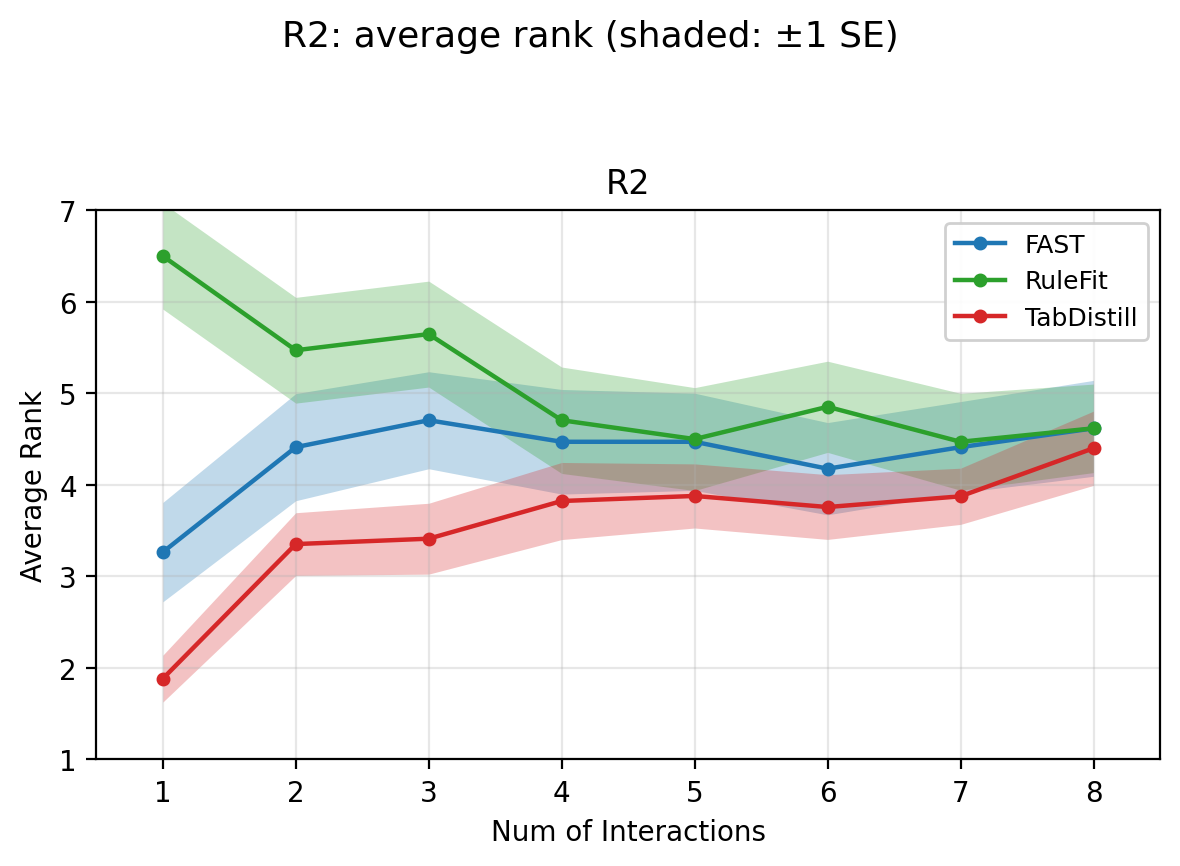}
    \caption{Regression tasks evaluated by $R^2$.}
    \label{fig:rank_reg_r2}
\end{subfigure}
\caption{\textit{Additional benchmark results in regression tasks.}}
\label{fig:appendix_reg_rank_comparison}
\end{figure*}

\begin{figure*}[t]
\centering
\begin{subfigure}[t]{0.48\textwidth}
    \centering
    \includegraphics[width=\linewidth,trim=0cm 0cm 0cm 2.5cm,clip]{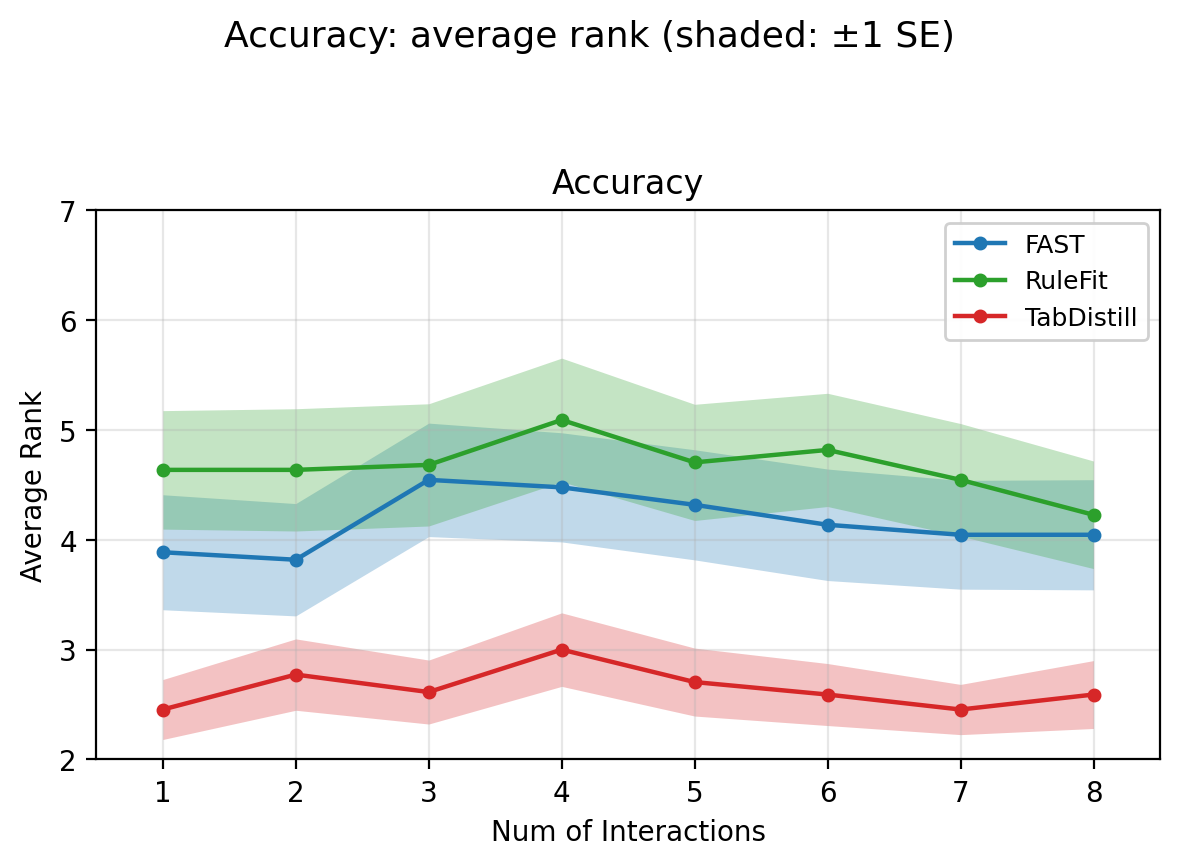}
    \caption{Classification tasks evaluated by Accuracy.}
    \label{fig:rank_cls_acc}
\end{subfigure}\hfill
\begin{subfigure}[t]{0.48\textwidth}
    \centering
    \includegraphics[width=\linewidth,trim=0cm 0cm 0cm 2.5cm,clip]{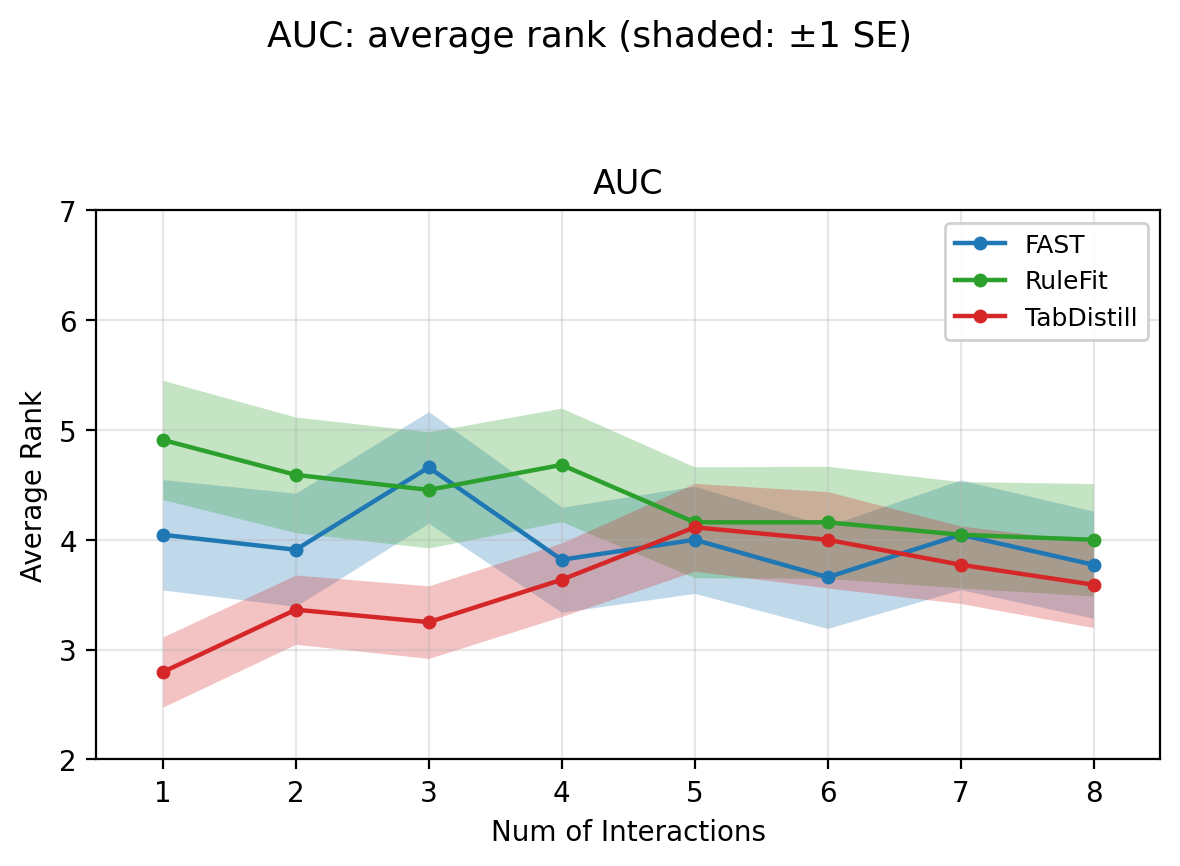}
    \caption{Classification tasks evaluated by AUROC.}
    \label{fig:rank_cls_auroc}
\end{subfigure}
\caption{\textit{Additional benchmark results in classification tasks.}}
\label{fig:appendix_cls_rank_comparison}
\end{figure*}

\subsection{Ablation Studies}\label{sec:ablation}

We next report additional ablation results for different interaction indices and different SPEX query budgets.

\begin{table}[H]
\centering
\small
\setlength{\tabcolsep}{6pt}
\renewcommand{\arraystretch}{1.15}
\label{tab:main_compare_index_reg}
\begin{tabular}{c c c c c c c c c c c}
\toprule
Metric & $N_{int}$ & \method & FAST & RuleFit & BII & Fourier & FSII & Mobius & SII & STII \\
\midrule
\multirow{8}{*}{MAE}
& 1 & 1.82 & 3.56 & 6.71 & 1.91 & 1.61 & 1.76 & \textbf{1.56} & 1.79 & 1.62 \\
& 2 & 3.00 & 4.68 & 5.94 & 2.79 & 3.09 & 3.00 & 2.91 & \textbf{2.74} & 3.00 \\
& 3 & 3.26 & 4.79 & 5.47 & 4.29 & 4.28 & \textbf{3.12} & 3.18 & 3.76 & 3.47 \\
& 4 & 3.62 & 4.18 & 5.21 & 5.24 & 4.97 & 3.62 & 3.65 & 4.91 & \textbf{3.44} \\
& 5 & \textbf{3.27} & 5.00 & 5.41 & 5.06 & 5.00 & 3.76 & 3.71 & 4.88 & 3.88 \\
& 6 & \textbf{3.27} & 4.68 & 5.15 & 6.10 & 5.07 & 3.55 & 4.27 & 5.44 & 3.48 \\
& 7 & \textbf{4.03} & 5.06 & 4.71 & 5.19 & 4.46 & 4.52 & 4.33 & 4.61 & 4.33 \\
& 8 & 4.23 & 4.91 & 4.71 & 4.60 & 5.68 & 4.19 & 4.48 & 4.20 & \textbf{4.06} \\
\midrule
\multirow{8}{*}{MSE}
& 1 & 1.88 & 3.26 & 6.50 & 1.74 & 1.55 & 1.82 & 2.00 & \textbf{1.47} & 2.12 \\
& 2 & 3.35 & 4.41 & 5.47 & 2.71 & \textbf{2.55} & 3.35 & 3.18 & 2.68 & 3.44 \\
& 3 & 3.41 & 4.71 & 5.65 & 4.29 & 4.16 & \textbf{3.09} & 3.15 & 3.67 & 3.50 \\
& 4 & 3.82 & 4.47 & 4.71 & 5.03 & 4.68 & \textbf{3.65} & 4.06 & 4.48 & 3.88 \\
& 5 & \textbf{3.88} & 4.47 & 4.50 & 4.70 & 5.16 & 4.09 & 4.15 & 4.59 & 4.44 \\
& 6 & 3.76 & 4.18 & 4.85 & 5.77 & 5.50 & 4.06 & 4.42 & 5.03 & \textbf{3.48} \\
& 7 & \textbf{3.88} & 4.41 & 4.47 & 5.23 & 5.12 & 4.85 & 4.76 & 4.61 & 4.09 \\
& 8 & 4.40 & 4.62 & 4.62 & 4.73 & 5.91 & 4.35 & 4.87 & 4.27 & \textbf{3.42} \\
\midrule
\multirow{8}{*}{$R^2$}
& 1 & 1.88 & 3.26 & 6.50 & 1.74 & 1.55 & 1.82 & 2.00 & \textbf{1.47} & 2.12 \\
& 2 & 3.35 & 4.41 & 5.47 & 2.71 & \textbf{2.55} & 3.35 & 3.18 & 2.68 & 3.44 \\
& 3 & 3.41 & 4.71 & 5.65 & 4.29 & 4.16 & \textbf{3.09} & 3.15 & 3.67 & 3.50 \\
& 4 & 3.82 & 4.47 & 4.71 & 5.03 & 4.68 & \textbf{3.65} & 4.06 & 4.48 & 3.88 \\
& 5 & \textbf{3.88} & 4.47 & 4.50 & 4.70 & 5.16 & 4.09 & 4.15 & 4.59 & 4.44 \\
& 6 & 3.76 & 4.18 & 4.85 & 5.77 & 5.50 & 4.06 & 4.42 & 5.03 & \textbf{3.48} \\
& 7 & \textbf{3.88} & 4.41 & 4.47 & 5.23 & 5.12 & 4.85 & 4.76 & 4.61 & 4.09 \\
& 8 & 4.40 & 4.62 & 4.62 & 4.73 & 5.91 & 4.35 & 4.87 & 4.27 & \textbf{3.42} \\
\bottomrule
\end{tabular}
\caption{
Rank comparison across different numbers of interactions for different \method index methods in regression tasks. 
Ranks are computed over the full set of models, including FAST and RuleFit. Lower is better.
}
\end{table}
\begin{table}[H]
\centering
\small
\setlength{\tabcolsep}{6pt}
\renewcommand{\arraystretch}{1.15}
\label{tab:main_compare_index_cls}
\begin{tabular}{c c c c c c c c c c c}
\toprule
Metric & $N_{int}$ & \method & FAST & RuleFit & BII & Fourier & FSII & Mobius & SII & STII \\
\midrule
\multirow{8}{*}{Accuracy}
& 1 & 2.45 & 3.89 & 4.64 & \textbf{2.02} & 2.52 & 2.64 & 2.66 & 2.34 & 2.27 \\
& 2 & 2.77 & 3.82 & 4.64 & 2.77 & 2.77 & 2.50 & 2.50 & \textbf{2.11} & 2.55 \\
& 3 & \textbf{2.61} & 4.55 & 4.68 & 3.09 & 2.88 & 2.75 & 2.64 & \textbf{2.61} & 2.75 \\
& 4 & 3.00 & 4.48 & 5.09 & 2.52 & 2.91 & 3.18 & \textbf{2.39} & 2.91 & 2.41 \\
& 5 & 2.70 & 4.32 & 4.70 & 2.98 & 3.60 & 2.68 & 2.61 & 2.75 & \textbf{2.45} \\
& 6 & 2.59 & 4.14 & 4.82 & 2.95 & 3.17 & 2.80 & 2.57 & 3.02 & \textbf{2.50} \\
& 7 & \textbf{2.45} & 4.05 & 4.55 & 2.77 & 3.17 & 2.70 & 3.18 & 2.89 & 2.95 \\
& 8 & \textbf{2.59} & 4.05 & 4.23 & 2.98 & 3.34 & 2.84 & 2.89 & 2.93 & 2.73 \\
\midrule
\multirow{8}{*}{F1 score}
& 1 & 2.77 & 3.66 & 4.43 & \textbf{2.23} & 2.89 & 2.95 & 3.00 & 2.75 & 2.57 \\
& 2 & 3.07 & 4.14 & 4.70 & 2.68 & 3.27 & 2.89 & 2.75 & \textbf{2.45} & 2.86 \\
& 3 & 2.91 & 4.39 & 4.61 & 3.25 & 3.40 & 2.93 & \textbf{2.77} & 2.98 & 2.98 \\
& 4 & 3.20 & 3.89 & 4.73 & \textbf{2.70} & 3.33 & 3.50 & 2.75 & 3.32 & 2.86 \\
& 5 & 3.05 & 3.98 & 4.27 & 3.02 & 3.81 & 3.14 & 3.30 & 3.07 & \textbf{2.91} \\
& 6 & \textbf{2.86} & 3.68 & 4.18 & 3.45 & 3.49 & 3.27 & 3.00 & 3.52 & 2.89 \\
& 7 & \textbf{2.82} & 3.66 & 4.11 & 3.26 & 3.55 & 3.14 & 3.36 & 3.41 & 3.20 \\
& 8 & \textbf{3.00} & 3.66 & 3.89 & 3.37 & 3.21 & 3.23 & 3.27 & 3.42 & 3.11 \\
\midrule
\multirow{8}{*}{AUROC}
& 1 & 2.80 & 4.05 & 4.91 & 2.52 & 3.18 & \textbf{2.41} & 2.80 & 2.80 & 2.59 \\
& 2 & 3.36 & 3.91 & 4.59 & 2.95 & 3.73 & 2.98 & 3.18 & \textbf{2.80} & 3.27 \\
& 3 & 3.25 & 4.66 & 4.45 & 3.30 & 4.07 & 3.30 & 3.18 & 3.41 & \textbf{2.95} \\
& 4 & 3.64 & 3.82 & 4.68 & 3.45 & 4.47 & 3.41 & \textbf{3.11} & 3.48 & 3.32 \\
& 5 & 4.11 & 4.00 & 4.16 & 3.91 & 4.71 & 3.50 & 3.50 & 3.52 & \textbf{3.11} \\
& 6 & 4.00 & 3.66 & 4.16 & 3.95 & 4.44 & 3.91 & \textbf{3.39} & 3.52 & 3.59 \\
& 7 & 3.77 & 4.05 & 4.05 & 3.88 & 4.33 & 4.02 & 3.68 & \textbf{3.36} & 3.84 \\
& 8 & \textbf{3.59} & 3.77 & 4.00 & 3.88 & 4.34 & 3.77 & 3.77 & 3.93 & 3.86 \\
\bottomrule
\end{tabular}
\caption{
Rank comparison across different numbers of interactions for different \method index methods in classification tasks. 
Ranks are computed over the full set of models, including FAST and RuleFit. Lower is better.
}
\end{table}

\begin{figure}[ht]
  \vskip 0.2in
  \begin{center}
    \centerline{\includegraphics[width=\columnwidth]{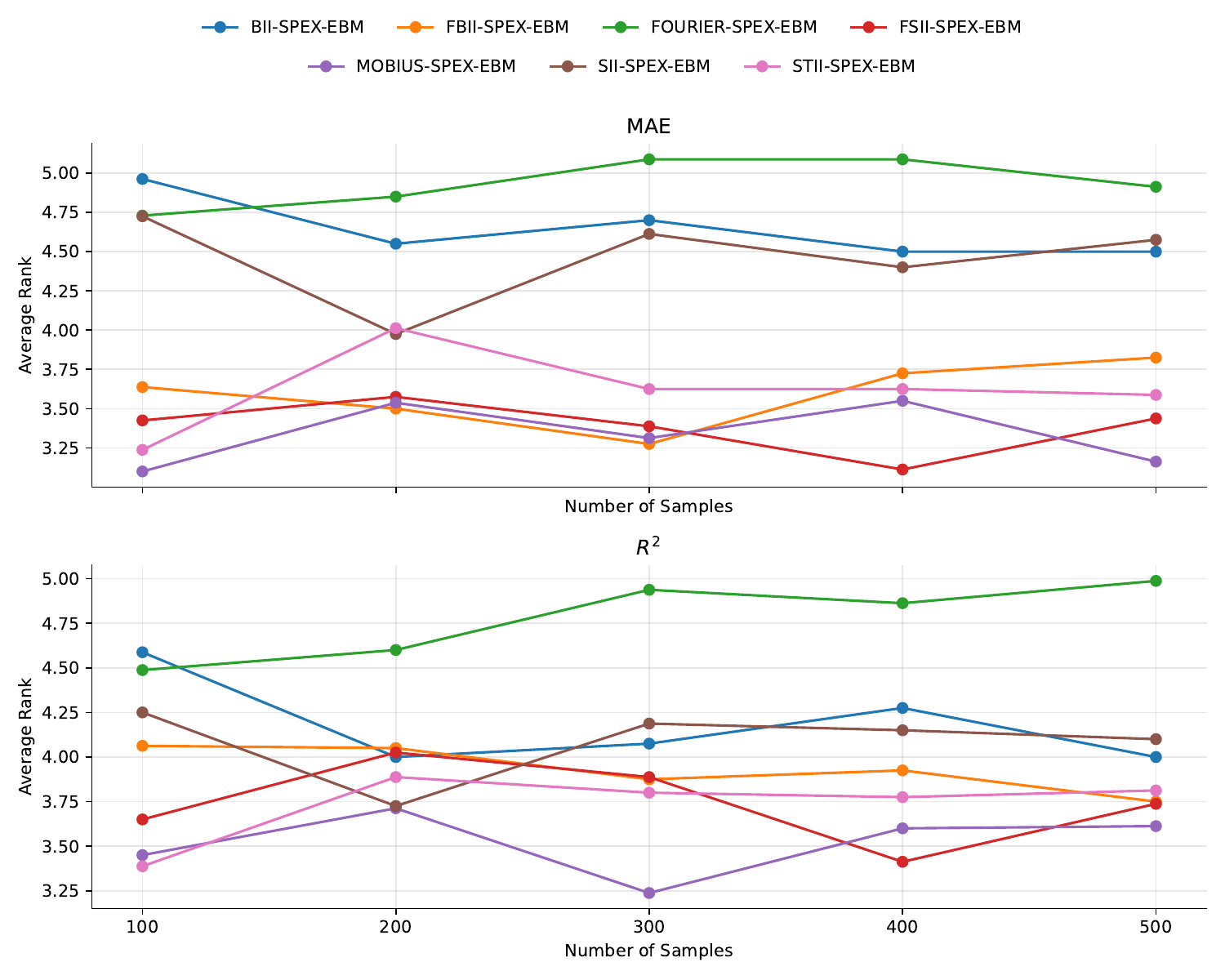}}
    \caption{
        \textit{Stability of interaction selection with respect to the SPEX query sample size.}
        We vary the query sample size used by SPEX from 100 to 500 and evaluate downstream predictive performance on TabArena regression datasets, reported as average rank across datasets. Across all indices, rank variations remain within approximately 0.5.
    }
    \label{fig:sample_size}
  \end{center}
\end{figure}

\clearpage

\subsection{Interaction Selection in the Fiat 500 Car Price Prediction Case Study}
\label{sec:case_study}

We provide additional details for the Fiat 500 price prediction case study. For each method, we select the top 10 interactions and train an EBM using the selected interaction set. \method captures higher-order interactions between geographic location and vehicle characteristics, including vehicle model and age.
In contrast, FAST is limited to pairwise effects and therefore captures only partial location-related interactions.
Although RuleFit can identify higher-order interactions, it fails to capture interactions involving the seller's geographic location in this task.

\begin{table}[H]
\centering
\small
\renewcommand{\arraystretch}{1.2}
\begin{tabular}{lcccc}
\toprule
\textbf{Method} & MSE & RMSE & MAE & $R^2$ \\
\midrule
\method &
509888 &
714.1 &
536.1 &
0.8595 \\
FAST &
540500 &
735.2 &
540.0 &
0.8511 \\
RuleFit &
523756 &
724.7 &
545.9 &
0.8557 \\
\bottomrule
\end{tabular}
\caption{Performance comparison for the Fiat 500 price prediction task.}
\label{tab:case_study}
\end{table}

\begin{table}[H]
\centering
\small
\setlength{\tabcolsep}{4pt}
\renewcommand{\arraystretch}{1.15}
\begin{tabularx}{\linewidth}{c >{\raggedright\arraybackslash}X >{\raggedright\arraybackslash}X >{\raggedright\arraybackslash}X}
\toprule
 & \textbf{TabDistill} & \textbf{FAST} & \textbf{RuleFit} \\
\midrule
1  & (vehicle model, vehicle age)
   & (vehicle model, vehicle age)
   & (vehicle age, mileage) \\
2  & (vehicle model, vehicle age, mileage)
   & (vehicle model, latitude)
   & (vehicle age, mileage, longitude) \\
3  & (vehicle age, mileage)
   & (vehicle model, longitude)
   & (mileage, longitude) \\
4  & (vehicle age, latitude)
   & (engine power, latitude)
   & (vehicle age, latitude) \\
5  & (vehicle model, vehicle age, latitude, longitude)
   & (vehicle age, mileage)
   & (vehicle age, mileage, latitude) \\
6  & (vehicle model, vehicle age, latitude)
   & (vehicle age, latitude)
   & (vehicle model, vehicle age, mileage) \\
7  & (vehicle model, mileage)
   & (vehicle age, longitude)
   & (vehicle model, vehicle age, mileage) \\
8  & (vehicle model, vehicle age, longitude)
   & (mileage, latitude)
   & (vehicle model, vehicle age, mileage) \\
9  & (vehicle model, latitude)
   & (mileage, longitude)
   & (vehicle model, vehicle age, mileage) \\
10 & (vehicle model, latitude, longitude)
   & (latitude, longitude)
   & (vehicle model, vehicle age, mileage) \\
\bottomrule
\end{tabularx}
\caption{
Comparison of selected feature interactions on the used car pricing task.}
\label{tab:case_study_interaction}
\end{table}

\begin{figure}[H]
\centering
\includegraphics[width=0.5\textwidth]{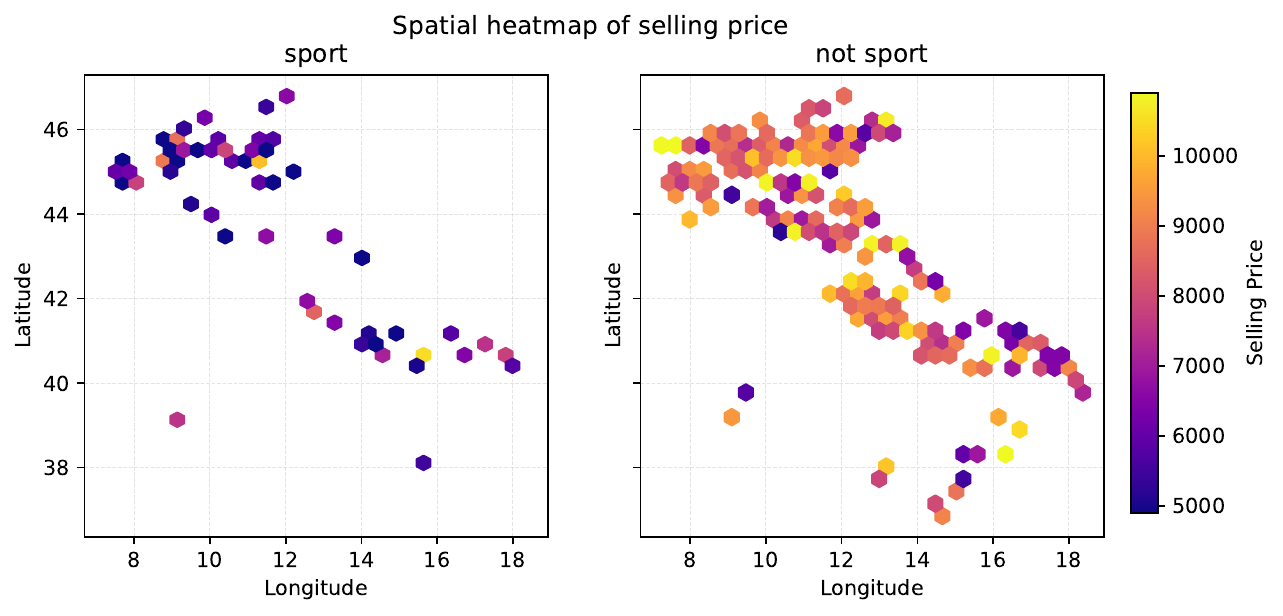}
\caption{\textit{A higher-order interaction identified by \method between seller location and model type.} The spatial pattern differs across model types. This indicates a non-additive interaction between longitude, latitude, and model type.}
\label{fig:case_study}
\end{figure}

\begin{table}[H]
\centering
\label{tab:model_location_median_price}
\begin{tabular}{lcc}
\toprule
& \multicolumn{2}{c}{Model type} \\
\cmidrule(lr){2-3}
Location cell & Sport & Non-sport \\
\midrule
South-West & 5.90 & 9.40 \\
South-East & 5.80 & 9.30 \\
North-West & 5.70 & 9.39 \\
North-East & 5.50 & 9.80 \\
\bottomrule
\end{tabular}
\caption{\textit{Median selling price (in \$k) by model type across \(2\times2\) location cells.} The four location cells are defined by splitting longitude and latitude at their median values.}
\end{table}